\documentclass[review, 3p]{elsarticle}

\usepackage[utf8]{inputenc}

\usepackage{amsmath, amssymb, bm}
\usepackage{amsmath}

\usepackage{booktabs}
\usepackage{multirow}

\usepackage{tikz}
\usetikzlibrary{arrows}

\usepackage{xspace}
\usepackage{xcolor}
\usepackage{lineno}

\usepackage{multirow}
\usepackage{tikz}

\usepackage{amsmath,amssymb,amsfonts}
\usepackage{algorithmic}
\usepackage{graphicx}
\usepackage{textcomp}
\usepackage{xcolor}
\usepackage{subcaption}

\usepackage[colorlinks, allcolors=black]{hyperref}
\usepackage{cleveref}

\usepackage{lineno,hyperref}

\journal{Journal of \LaTeX\ Templates}

\usepackage{pdfpages}

\begin{document}

\begin{frontmatter}

\title{Synthesis of 3D On-Air Signatures with the Sigma-Lognormal Model}

\author[ulpgc]{Miguel A. Ferrer}
\author[ulpgc]{Moises Diaz\corref{mycorrespondingauthor}}
\cortext[mycorrespondingauthor]{Corresponding author}
\ead{moises.diaz@ulpgc.es}
\author[ulpgc]{Cristina Carmona-Duarte}
\author[ulpgc]{Jose Juan Quintana}
\author[epm]{Réjean~Plamondon}

\address[ulpgc]{Instituto Universitario para el Desarrollo Tecnológico y la Innovación en Comunicaciones (iDeTIC), Universidad de Las Palmas de Gran Canaria, Spain. Emails: \{miguelangel.ferrer, moises.diaz,cristina.camona,josejuan.quintana\}@ulpgc.es}
\address[epm]{École Polytechnique, University of Montreal, Montreal, QC, Canada.  Email: rejean.plamondon@polymtl.ca}

\begin{abstract} 
Signature synthesis is a computation technique that generates artificial specimens which can support decision making in automatic signature verification. A lot of work has been dedicated to this subject, which centres on synthesizing dynamic and static two-dimensional handwriting on canvas. This paper proposes a framework to generate synthetic 3D on-air signatures exploiting the lognormality principle, which mimics the complex neuromotor control processes at play as the fingertip moves. Addressing the usual cases involving the development of artificial individuals and duplicated samples, this paper contributes to the synthesis of: (1) the trajectory and velocity of entirely 3D new signatures; (2) kinematic information when only the 3D trajectory of the signature is known, and (3) duplicate samples of 3D real signatures. Validation was conducted by generating synthetic 3D signature databases mimicking real ones and showing that automatic signature verifications of genuine and skilled forgeries report performances similar to those of real and synthetic databases. We also observed that training 3D automatic signature verifiers with duplicates can reduce errors. We further demonstrated that our proposal is also valid for synthesizing 3D air writing and gestures. Finally, a perception test confirmed the human likeness of the generated specimens. The synthesizer and all the databases generated are publicly available, only for research purposes, at {\it to be added after the acceptance of the paper.}
\end{abstract}


\end{frontmatter}

 
\section{Introduction}

This work was carried out in the context of active 3D signature synthesis research. A 3D signature can be defined as a rapid and overlearned complex movement of the fingertip in the air. Because this movement can be considered as the final response of a set of complex neuromotor systems, computational models inspired by biological principles can be used to synthesize 3D signatures.  

\textcolor{black}{Many human-related computer vision problems can be approached by analyzing the kinematics of human movements~\cite{winter2009biomechanics}. Kinematics variables include linear and angular displacements, velocities, and accelerations. These data typically register the kinematics of anatomical landmarks such as the center of gravity of body segments, the centers of joint rotations, the end effectors of limb segments, key anatomical prominences, etc. }

\textcolor{black}{Exploring natural human kinematics provides fundamental insight allowing to understand the mechanisms that characterize natural human motion, the higher level strategies of realizing complex tasks, and the interactions with the external environment. It might be useful to develop tools for characterizing human motion changes due to disease, aging, or injury. This can help in understanding how fine motor control is learned by elementary school children, and can provide quantitative and objective learning indexes. It can facilitate rehabilitation applications, including the design of prostheses and exoskeletons, workplace ergonomics, sports medicine, orthopedics, and physical therapy, which are of paramount importance for human welfare. }

\textcolor{black}{Furthermore, the synthesis of such kinematics opens wide the door to predicting movements and synthesizing human-like motions and developing strategies for human motion reconstruction on engineered anthropomorphic systems, such as humanoids, mobile manipulators and simulated systems.}

Many theories have attempted to approach the kinematics of human movement in general~\cite{plamondon2014recent}.  Among these, the Kinematic Theory of Rapid Human Movements and its associated Delta-Lognormal  \cite{plamondon1998generation} and Sigma-Lognormal $(\Sigma\Lambda)$ \cite{o2009development} models have been extensively used to explain most of the fundamental phenomena covered by classical studies on human motor control and to study several factors involved in fine motricity.

\subsection{Related Work}

 \textcolor{black}{The synthesis of the human-like kinematics of 3D movements has emerged as an alternative to the limitations imposed by using real data. Certainly, freely using, sharing, or processing sensitive human data advances technological development. However, this is hindered by data regulations and laws, such as the European General Data Protection Regulation. Moreover, acceptance of synthetic data has increased with enhanced realism, as presented in the literature. Indeed, it is often hard to differentiate such data from real data relating to humans and machines.}
 
\textcolor{black}{The synthesis of kinematics is especially suited to computer-based applications. For example, computer systems are starting to replace real data with synthetic specimens in a bid to reduce privacy concerns. Also, this synthesis can balance data in training or augment it through duplication strategies. Synthesis can also be applied to computational classifiers to optimize initializing model weights or fine-tuning features. Other areas focus on improving the realism of synthetic samples. Furthermore, the generation of large databases has allowed both industry and academia to evaluate algorithms under common benchmarks~\cite{joshi2022synthetic}.}
 
 \textcolor{black}{These domains of application also allow facing several human analyses assisted with synthetic data; for example, the estimation of a 3D head pose~\cite{basak2021learning}, the recovery of a skeleton-based body pose~\cite{menier20063d}, improvements in smartphone user authentication~\cite{buriro2021swipegan} or action recognitions through the use of phones with the hands~\cite{hatay2021learning}, etc.}

 Over the last decade, comprehensive research on the $\Sigma\Lambda$ model has been conducted to generate complete databases of synthetic signatures in 2D \cite{ferrer2017static, galbally2012synthetic, diaz2019perspective}. Initially, a master signature with its $\Sigma\Lambda$ paremeters is designed. Then, repetitions of this master signature are generated by randomly modifying their parameters. Similarly, skilled forgeries for synthetic genuine users are also generated \cite{ferrer2019generating, ferrer2016behavioral, lai2021synsig2vec}.

 Furthermore, given a spatial trajectory of a signature, some proposals have been oriented to add the signature's kinematic properties with the  $\Sigma\Lambda$ model. This addition has led to investigations of how to recover temporal data~\cite{diaz2017recovering}, make a signature more fluent~\cite{diaz2021improving}, or render a fake specimen more skilfully~\cite{ferrer201improving}. 
  
  One advantage of modelling a real dynamic signature with the $\Sigma\Lambda$ is the possibility of generating duplicates by modifying the signature's parameters~\cite{galbally2012synthetic}.Combining real enrolled signatures with such duplicates has improved the performance of Automatic Signature Verifiers (ASVs)~\cite{diaz2016dynamic}. Such a combination is, however, challenging since the training includes real and synthetic signatures. Similar strategies have also been applied to generate handwriting~\cite{carmona2017temporal}, duplicate the generation of handwriting~\cite{fischer2014neuromuscular} or, to synthesize gestures~\cite{leiva2015gestures}.

 The above synthesis of 2D signatures has evolved to the development of a new trend in which on-air signatures are synthesized in 3D. The growth of motion sensors such as the Leap Motion, the creative senz3D camera, inertial measurement units, etc., has led to the acquisition of new 3D on-air signature databases. The latter include several users with genuine and skilled forgeries repetition such as the DeepAirSig~\cite{malik2020deepairsig}, 3DIIT Signatures~\cite{behera2018analysis}, or other in-house datasets~\cite{guerra2021air}. Experiments verifying 3D signatures of these databases~\cite{guerra2021air} suggest that the third dimension offers an extra security layer. Based on the lesson learnt in synthesizing 2D signatures, synthetic signatures in 3D can impulse the development of new automatic signature verifiers and improve the performance of the systems. Some challenges resolved through the synthesis include sharing 3D data without being subject to legal procedures and data protection law, conducting scalable data analysis, improving the robustness of biometric systems against attacks~\cite{ferrer2018biometric}, and designing databases to avoid recruiting volunteers and eliminating human errors in labelling specimens, among others. In sum, all the advantages of synthesis of signatures in 2D will be directly transferred to 3D.

\subsection{Our Contributions}

\begin{figure}
\centering
 \includegraphics[width=0.9\linewidth]{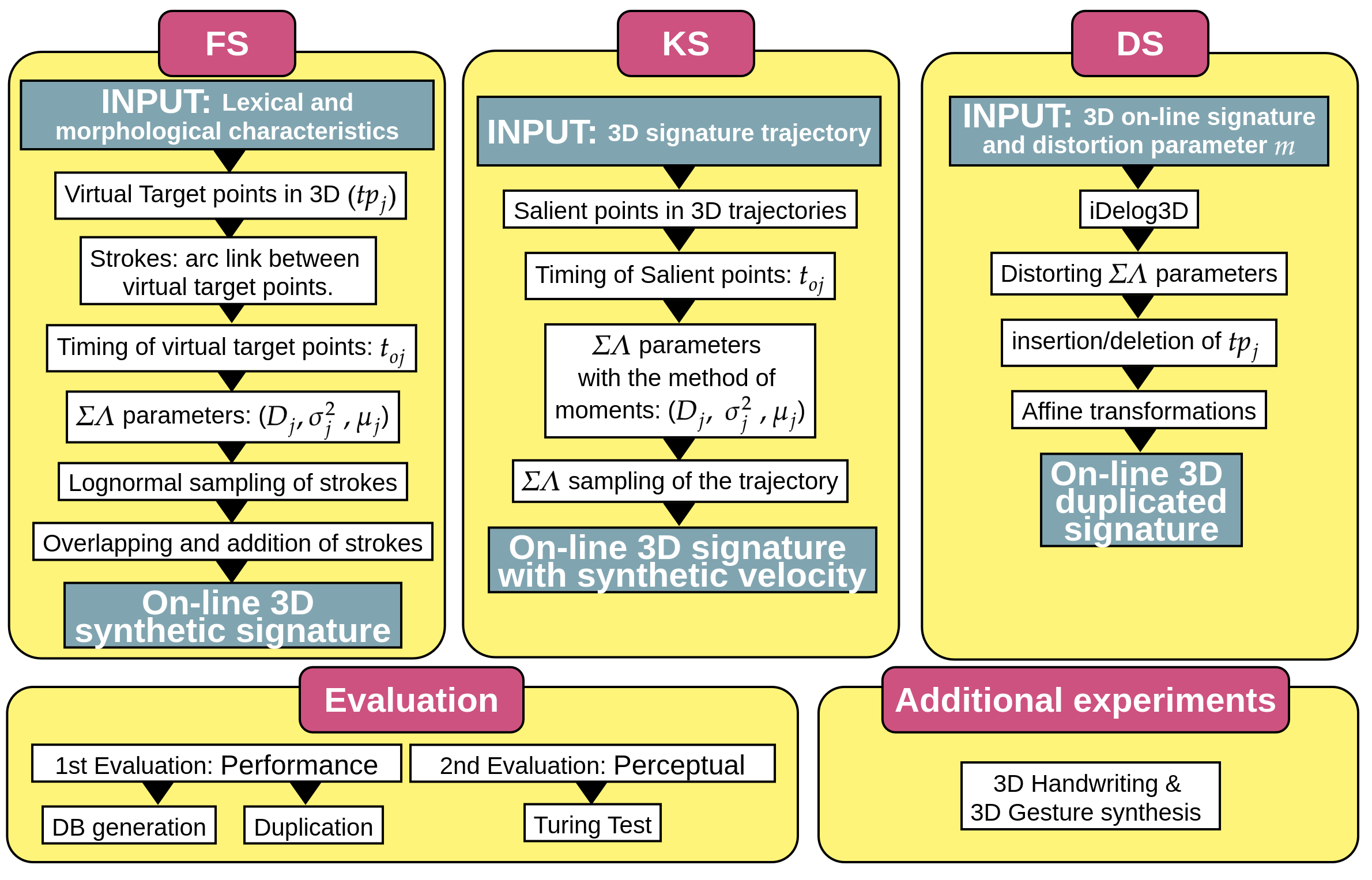}
  \caption{Graphical diagram of the contributions of this article: three methods for designing 3D signatures, their evaluation, and additional experiments.}
  \label{fig:diagram} 
\end{figure}

 In this paper, the main proposal is a procedure to generate synthetic 3D signatures. The procedure is extended to create databases of such signatures. Specifically, we consider five innovations to the state of the art:

  \begin{enumerate}
   \item     We synthesize entirely 3D on-air signatures belonging to synthetic identities. This is done by reversing an implementation of the Sigma-Lognormal model in 3D called iDeLog3D~\cite{idelog3D_2022}. This method is known as a Full Synthesis (FS).
    
    \item Next, we provide a method to generate human-like velocity profiles in 3D signatures when only the spatial 3D trajectory is available without time information. In other words, only the 3D trajectory position is used as input. This method is known as Kinematic Synthesis (KS).  
    
    \item Duplicated on-air signatures of real specimens are also generated. The same duplication procedure synthesizes 3D genuine and skilled forgeries. This method is named Duplicated Synthesis (DS).
    
    \item To demonstrate the flexibility of our model, we generate the synthetic counterparts of several publicly available 3D signature databases. 
    
    \item We conduct extensive experiments demonstrating that real and synthetic signature databases have similar quantitative and subjective performances. A Turing test is conducted to judge the human likeness of the synthetic samples through videos of signatures. Furthermore, we investigate the capacity of our synthesizer to generate air writing and gestures in 3D, obtaining competitive results as well.
  \end{enumerate}

    To the best of our knowledge, this work is the first to provide open on-air synthetic signatures. A graphical diagram of the contributions of the present article and the evaluations of the proposed methodologies are illustrated in Figure~\ref{fig:diagram}. The generated databases and synthesizers are publicly available only for research at {\it to be added after the acceptance of the paper}.

 The remainder of the paper is organized as follows: Section II presents our contribution to the synthesis of 3D on-air signatures of synthetic identities (FS method), while the generation of the kinematics of a given 3D signature trajectory (KS method) is explained in Section III. Section IV provides our contribution to duplicated genuine and skilled forgeries (DS method). An extensive experimental validation of the synthesis is reported in Section V. Finally, Section VI concludes \textcolor{black}{and discusses} the article by summarizing its key contributions.

\section{FS: Full-based Synthesis of 3D Signatures}

 The FS procedure proposes a method for synthesizing the trajectory and kinematics of human-like 3D on-air signatures\footnote{Note that for kinematic synthetizes, this work refers to the speed profile of the fingertip as the single end effector.}. \textcolor{black}{This procedure allows for data replacement experiments in signature verification, avoiding data protection issues. The FS also helps in setting up standard benchmarks and massive ASV evaluations.}

 First, we define a set of virtual target points for the trajectory and the link between them. Then, for velocity, we assign a reaching time to each virtual target point and a lognormal sampling of each link between virtual target points. Finally, a summation of all the time-overlapped links between virtual target points generates the 3D signature. As a result, a master signature of the synthetic identity is obtained.

 \subsection{Synthesis of Virtual Target Points and Links in 3D}
 
  According to kinematic theory, a signature trajectory can be designed through a set of virtual target points, $tp_j$,  connected by circular arcs of circumference. These points represent the cognitive action plan for a human movement~\cite{plamondon2021lognormality}. In addition, to synthesize credible signatures from scratch, a language model may be of help. Such a model would produce signatures with (semi-)legible names, flourishes, or a combination of name and flourishes.
 
 Similar to our previous works~\cite{ferrer2017static, ferrer2016behavioral, ferrer2014static}, we used the lexical and morphological characteristics of real signatures~\cite{diaz2015modeling} in this article. They provided the distribution of characters and flourishes in terms of virtual target points in 2D. Then, we connected them with an arc of circumference.
 
 The $tp_j$ and the midpoints of the circular arcs, denoted $mp_j$, are projected to 3D through the following sinusoidal 3D surface: $z = A_x \sin(\omega_x x + \phi_x) + A_y \sin(\omega_y y + \phi_y)$, where $(x, y)$ are the coordinates of the defined 2D $tp_j$ and $mp_j$. The surface parameters $(A_x, A_y, \omega_x, \omega_y, \phi_x, \phi_y)$ were set up heuristically in a trial-and-error fashion to obtain a similar appearance between the real and synthetic specimens. 
 
\begin{figure}
\centering
 \includegraphics[width=0.66\linewidth]{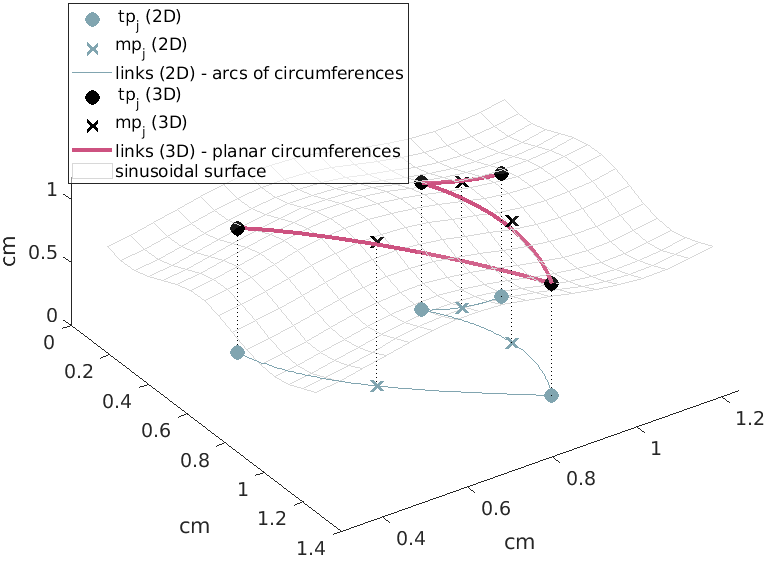}
  \caption{Generation of the action plan of letter ``v'' in 3D from the generated virtual target points and links in 2D through their projected points on a sinusoidal surface. The following abbreviations are used in the figure: $vp_j$ (blue) -  the generated virtual target point in 2D; $mp_j$ (blue) - the middle target point in 2D; the planar links (blue) connecting two consecutive virtual target points in 2D;  $vp_j$ (black) - the projection of the 2D virtual target points onto a 3D sinusoidal surface; $mp_j$ (black) - the 2D middle points projected onto the sinusoidal surface, and 3D planar links (magenta) between virtual target points.}
  \label{fig:FS} 
\end{figure}

 Finally, using the tuple $(tp_j, tp_{j+1}, mp_j)$ we defined a plane in 3D. Next, a planar circumference, passing through these three points in 3D, was generated for each target point. This procedure is visually exemplified in Figure~\ref{fig:FS} with the letter ``v''. Firstly, the virtual target points and circular arcs of the letter were generated in 2D, and then later projected onto 3D.  
 
 \subsection{Synthesis of Velocity}
   
   The timing of the virtual target points and the lognormal velocity function parameters are required to generate the 3D on-air signature.
   
   \subsubsection{Timing of virtual target points:}
   
   The virtual target points are timed following a periodic pattern based on the so-called Central Pattern Generators (CPG), which produce rhythmic patterned outputs to activate different motor pools~\cite{garipelli2006neuromotor}. This can be observed in the periodic pattern of human kinematics. Thus, a timestamp is assigned to each virtual target point as  $ts_j = ts_{j-1} + r, \forall j=1,\ldots, Ns$, where $ts_0$ is the beginning of the movement, and $r$ is a random value which follows a normal distribution $\mathcal{N}(0.1; 0.005)$ \cite{ferrer2017static}.

   \subsubsection{Velocity parameters:}
   
   The velocity profile, $v_j(t)$, of the links between the virtual target points $tp_{j-1}$ and $tp_j$ is defined with a family of lognormal equations as follows:
   
   \begin{equation}
    v_j(t) = \dfrac{D_j}{\sigma_j \sqrt{2\pi}(t-t_{0,j})} \text{exp}\left\lbrace\dfrac{-\left[ \text{ln}(t - t_{0,j}) - \mu_j\right]^2}{2\sigma_j^2} \right\rbrace
   \end{equation}
   
   where $t$ is the basis of time, $t_0$ is the time of stroke occurrence, $D_j$ is the amplitude of the input commands, $\mu_j$ is the stroke time delay on a logarithmic time scale and  $\sigma_j$ is the stroke response time. The values of $D_j, \mu_j$  and $\sigma_j^2$ are based on the experimental observations showing that the velocity profile starts to increase at $tp_{j-1}$, reaches a maximum around the centre of the stroke, and decreases up to $tp_j$. Additionally, the adjacent strokes are temporally overlapped.

   There are many solutions to $t_{0,j}, D_j, \mu_j$  and $\sigma_j^2$ fits the bell-shaped speed profile of a stroke~\cite{diaz2021improving}. The solution adopted heuristically in this paper is to select $t_{0,j} = ts_{j-1} - 0.5$~\cite{ferrer2018idelog}, giving enough time to the lognormal function to increase at the beginning of the stroke in $ts_{j-1}$. The parameter $D_j$ is computed as the distance traveled from $tp_{j-1}$ to $tp_{j}$.

   To calculate $\mu_j$ and $\sigma^2_j$, we build a system of two equations with these two unknowns. The first equation is obtained, bearing in mind that the distance $D_j$ traveled from  $tp_{j-1}$ to $tp_j$ is equal to the integral of the stroke velocity profile of the 3D Sigma-Lognormal model. Hence, it holds that:
   
   \begin{equation}
    D_j \approx \displaystyle \int_{t_{0,j}}^{ts_j} D_j v_j(t)\,dt = \dfrac{D_j}{2} \left\lbrace 1 + \text{erf} \left[ \dfrac{\text{ln}(ts_j - t_{0,j}) - \mu_j}{\sqrt{2}\sigma_j} \right]  \right\rbrace
   \end{equation}

   \noindent which calls the first equation of the system, given by:
   
   \begin{equation}\label{eq2}
    \dfrac{\text{ln}(ts_j - t_{0,j}) - \mu_j}{\sqrt{2} \sigma_j} = 3 \rightarrow \mu_j = \text{ln}(ts_j - t_{0,j}) - 3 \sqrt{2} \sigma_j
   \end{equation}
   
   \noindent since, by definition, from a practical point of view, $\text{erf}(3) = 1$. 
   
   Then, we consider the experimental fact that the lognormals are centred around the middle of the stroke. Because we assume that the peak of the lognormal occurs at $e^{\mu_j - \sigma^2_j}$ and the centre of the stroke occurs at time $(ts_{j-1} + tp_j)/2$, the second equation of the system is given by:
   
   \begin{equation}\label{eq3}
    (ts_{j-1} + tp_j)/2 - t_{0,j} = e^{\mu_j - \sigma^2_j} 
   \end{equation}

 \noindent The parameter $\sigma^2_j$ is obtained by solving the second-degree equation obtained when  \eqref{eq2}  and \eqref{eq3} are combined:
   
   \begin{equation}
    \sigma^2_j + 3\sqrt{2} \sigma_j  - \text{ln}(3/2) = 0
   \end{equation}

   \noindent and then $\mu_j$ can be drawn by substituting the $\sigma^2_j$ value in \eqref{eq2}.
   
   \subsection{3D On-air Signature Reconstruction}

   Once the velocity profiles $v_j(t)$ of each link between virtual target points are established, each of the links, defined by arcs of circumferences, is sampled at $f_m$ amples per second. The distance from the virtual target point $tp_{j-1}$ to the $k^{th}$  8-connected sample of the circular link between $tp_{j-1}$ and $tp_j$ is given by:
   
   \begin{equation}
    d_{k,j} \approx \displaystyle \int_0^{k/f_m} D_j v_j(t)\,dt, \qquad \forall j=1,\ldots,N, \qquad k=0,1,\ldots, f_m\cdot T
   \end{equation}

 \noindent  From $d_{k,j}$, the coordinates of the links between virtual target points $\left(x_j(t), y_j(t), z_j(t)\right)$ can be obtained and the synthetic trajectory reconstructed as follows:

   \begin{equation}
    \left(x_r(t), y_r(t), z_r(t)\right) = \displaystyle \sum_{j=1}^{tp_n} \left(x_j(t), y_j(t), z_j(t)\right)
   \end{equation}

\begin{figure} 
\captionsetup[subfigure]{labelformat=empty}
    \centering
  \subfloat[\label{FS11a}]{%
       \includegraphics[width=0.5\linewidth]{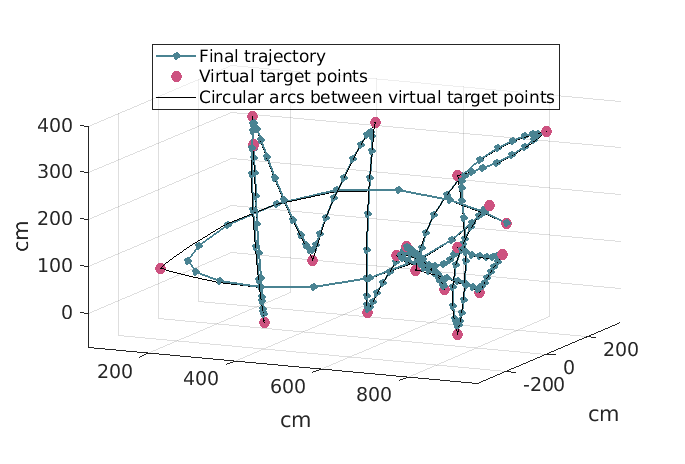}}
    \hfill
  \subfloat[\label{FS11b}]{%
        \includegraphics[width=0.5\linewidth]{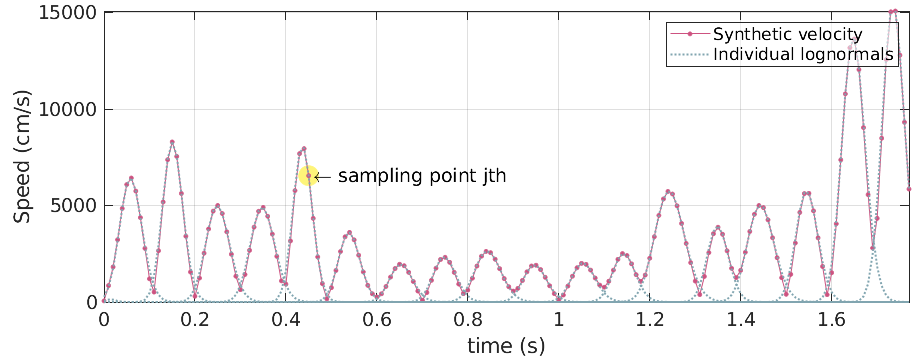}} 
  \caption{Example of a master 3D synthetic signature obtained with the FS method.}  
  \label{fig:FS1} 
\end{figure}

   Figure~\ref{fig:FS1} shows an example of a master synthetic signature (trajectory and velocity) obtained from a set of virtual target points randomly generated from the lexical and morphological characteristics. 
   
\section{KS: Synthesis of the Kinematics for a Signature Trajectory}

 This section synthesizes the kinematics of a given 3D trajectory as a sequence of ordered points. \textcolor{black}{This method can be applied to stabilize the kinematics of deficient acquisition devices because they register data with unstable frequency rates,  or jitter in the timing signal, among other noises. Indeed, synthetic kinematics can boost the performance of real kinematics in signature verification~\cite{diaz2015towards}. Note that many recognizers perform classifications according to the kinematic properties of signatures. Another case that KS can solve is that of using kinematics as a predictor, which may improve the quality of the acquired samples. Also, kinematic estimation is typically the final stage in a whole off-2-on challenge in 2D. In a follow-up work~\cite{diaz2021improving}, KS is used to adapt our previous work to the case of 3D signatures.}
 


 Firstly, KS determines the salient points which are directly related to the target points~\cite{plamondon2021lognormality}. Next, a velocity profile is assigned between two consecutive salient points. Finally, the 8-connected trajectory is sampled according to the estimated velocity. 

\subsection{Salient Points Estimation}

 The salient points are defined in an 8-connected 3D trajectory as the points with maximum curvature. Obtaining the curvature from the 3D trajectory misses some relevant salient points. As a result, we worked out the curvature in the three planes, $xy, xz$, and $yz$. Then, we combined the resulting salient points. Thus, the curvature in the plane $xy$ is calculated as:

\begin{equation}
   \kappa(x,y) = \dfrac{\sqrt{ \left(\ddot{y}\dot{x} - \ddot{x}\dot{y} \right)^2}}{\sqrt{\left( \dot{x}^2 + \dot{y}^2 \right)^{3}}} 
\end{equation}

 The salient points in each plane are obtained as follows~\cite{diaz2021improving, ferrer2017two}: he curvatures at each plane are worked out with a multiscale algorithm~\cite{ferrer2017two} at dozen scales, uniformly distributed between $1$ and $M/2$, with $M$ being the length of the trajectory, and the scale, the number of points used to work out the derivate. All the curvatures calculated are added together to obtain the summed curvature $C$. The salient points at each plane are selected as the peaks of $C$ whose height/width ratio are greater than $\left(\max(C) - \min(C) \right)/45$. The salient points of the 3D trajectory are obtained as the union of the salient point of the $xy, xz$,  and $yz$ planes.

\subsection{Synthesis of Velocity}

 Given the salient points, the velocity synthesis is based on a sum of lognormal functions inserted between consecutive salient points. There are many solutions to $t_{0,j}, D_j, \mu_j$   and $\sigma^2_j$ fits the bell-shaped speed profile of a stroke~\cite{diaz2021improving}. According to the synergy of the neighbouring lognormals considered by the kinematic theory, an inevitable overlapping among adjacent lognormals is considered.

 Initially, a time is assigned to each salient point, $sp_j$. This time is roughly available in the case of a low or unstable sampling frequency. If we do not rely on this timing, the salient points can be timed as in the FS method. We heuristically chose the following relation once again: $t_{0,j} = ts_{j-1} - 0.5$, giving enough time to the lognormal function to increase at the beginning of the stroke in $ts_{j-1}$.

 To calculate the parameters $\mu_j$ and $\sigma^2_j$, we define a bell shape with a mean $M_j$ equal to the centre $ts_{j-1} + (ts_j  - ts_{j-1})/2$ and variance $V_j = (ts_j - ts_{j-1})/4$ (it takes 95.45\% of energy in the segment and the rest of the overlap). This is an experimental value based on the general observations in the databases used in this article. The lognormal parameters are then recovered by the method of moments~\cite{bowman2004estimation}:

\begin{equation}
 \mu_j = \text{log}\left(M_j/\sqrt{V_j+M^2_j} \right); \qquad \sigma^2_j = \text{log}\left(V_j/M^2_j+1 \right)
\end{equation}

 \noindent And $l_{s,j} = D_j$, where $l_{s,j}$ is the distance between $sp_{j-1}$ and $sp_{j}$ in the 8-connected 3D trajectory. Therefore, the velocity module can be defined as: 

\begin{equation}\label{eq10}
 v(t) = \omega \displaystyle \sum_{j=1}^{Nblog} D_j v_j(t)
\end{equation}

\noindent where $Nblog$ is the number of lognormals and $\omega$ a scalar that adjusts the final area of the velocity to the distance travelled in the trajectory because overlap inaccuracies exist between lognormals as: $\omega = Ls / \int_0^T v(t)\,dt$, where $Ls$ is the length of the trajectory and $T$  the signature duration.

\subsection{Final Signature}

 Finally, the 3D 8-connected trajectory is sampled using the velocity module of Eq.~\eqref{eq10}. To this end, a frequency of the output signal is assigned to mimic the samples per second, $f_m$, typically defined by the acquisition device. Let $k$ be a sampling point in the online 3D signatures, and $k/f_m$ the timestamp of such sampling points, then the distance $d_k$ between samples in the 8-connected trajectory can be defined as follows:

\begin{equation}
 d_k = \displaystyle \int_0^{k/f_m} v(t)\,dt
\end{equation}

 This distance, $d_k$, is later used to select the samples that satisfy the defined velocity profile, obtaining the online signature $\left(x(t), y(t), z(t)\right)$.

\begin{figure} 
\captionsetup[subfigure]{labelformat=empty}
    \centering
  \subfloat[\label{KS1a}]{%
       \includegraphics[width=0.5\linewidth]{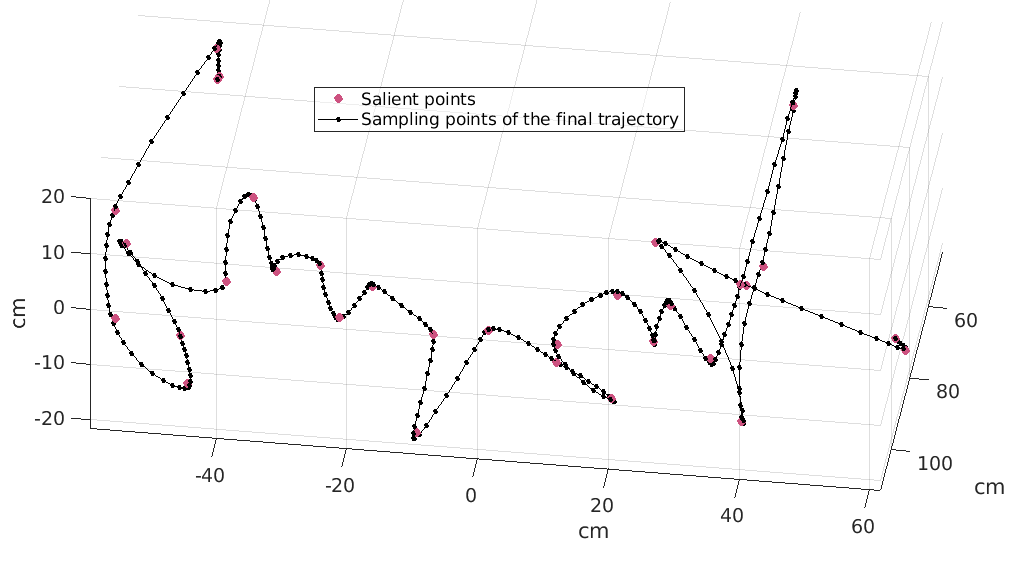}}
    \hfill
  \subfloat[\label{KS1b}]{%
        \includegraphics[width=0.5\linewidth]{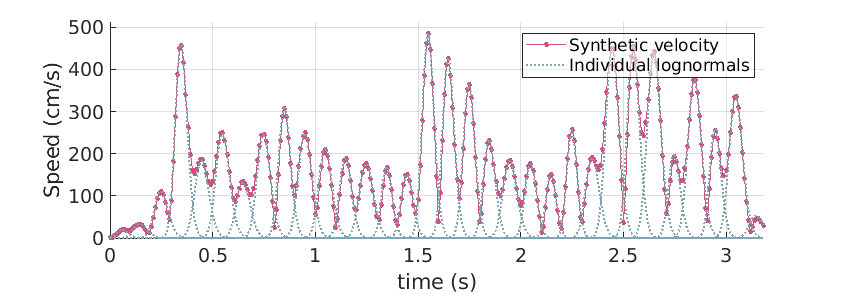}} 
  \caption{Example of a 3D signature with a sampled trajectory and a synthetic velocity through the KS method.}  
  \label{fig:KS} 
\end{figure}

 A visual example is illustrated in Figure~\ref{fig:KS}. Note that the trajectory is given as input, whereas the KS method synthesized the velocity.

 \section{DS: Duplicated Synthesis of 3D Real Signatures}

 It is well-known that two 2D signatures of a signer are never equal. The differences here constitute what is called intra-person variability. In 3D on-air signatures, these differences are even more evident. 3D movement is carried out by hands and arms, and this does not allow as fine a control as that of finger movements. A greater intra-person variability by duplication of signatures is thus expected in 3D. 

 A duplicated specimen is generated from a given online signature, which can be a real signature or a synthesized one. As well, a skilled forgery can be seen as a sort of duplicated signature with a different intra-personal variability.

 In our duplicated synthesis (DS), an online signature is analysed and parametrised with iDeLog3D~\cite{idelog3D_2022}. Then, the signature is distorted by defining a global parameter and modifying the value of the iDeLog3D parameters.

 \textcolor{black}{The DS method can help train signature verifiers. For example, it can balance the number of signatures required during training. Also, our method can augment the training set. Likely, this latter application is the most popular in the literature~\cite{galbally09ICDARenrollment}.}
 
 First, we define the parameter $m\in (0,1)$, which controls the gradual deformation of the movement. The higher the $m$ value, the greater the changes in the trajectory and velocity. Then, the iDeLog3D parameters are distorted by applying the following transformations:

\begin{equation}
 \begin{split}
\mu'_j & = \mu_j (1+0.01\cdot m\cdot  \mathcal{N}(0; Nblog) )\\
\sigma_j^{2'} & = \sigma^2_j (1+0.01\cdot m\cdot  \mathcal{N}(0; Nblog) )\\
t'_{0,j} & = t_{0,j}  (1+0.001\cdot m\cdot  \mathcal{N}(0; Nblog) )\\
\vartheta'_{e,j} & = \vartheta_{e,j}  (1+0.001\cdot m\cdot  \mathcal{N}(0; Nblog) )\\
\vartheta'_{s,j} & = \vartheta_{s,j}  (1+0.001\cdot m\cdot  \mathcal{N}(0; Nblog) )\\
\phi'_{e,j} & = \phi_{e,j}  (1+0.001\cdot m\cdot  \mathcal{N}(0; Nblog) )\\
\phi'_{s,j} & = \phi_{s,j}  (1+0.001\cdot m\cdot  \mathcal{N}(0; Nblog) )
\end{split}
\end{equation}

 Next, the virtual target points are altered. Let $N$ be the number of virtual target points calculated in the master 3D signature and $N'$ the points in a duplicated signature. We randomly remove the closest salient points between 0\% and 5\%. Similarly, new points are added among the 0\%-5\% of the nearest point. As a result, we have a new  $tp_j, \forall j=1, \ldots, N'$ sequence. Inserting and removing target points affects the fluency of the signature~\cite{diaz2021improving}. Typically, a duplicated skilled forgery requires adding more $tp_j$. 

 In 3D, the positions of virtual target points are defined as: $tp_j = \left( tpx_j, tpy_j, tpz_j \right)$. Similar to our previous work designing duplicates~\cite{diaz2016dynamic}, the points' position is modified with non-linear distortions as follows:

\begin{equation}
 \begin{split}
tpx'_j & = tpx_j \left(1+A_x \sin(2\pi P_x tpx_j/L_x) \right)\\
tpy'_j & = tpy_j \left(1+A_y \sin(2\pi P_y tpy_j/L_y) \right)\\
tpz'_j & = tpz_j \left(1+A_z \sin(2\pi P_z tpz_j/L_z) \right)
\end{split}
\end{equation}
 
 \noindent  where $(L_x, L_y, L_z)$ are the width and height of the signature defined with the virtual target points margins, $(A_x, A_y, A_z) = m/50 (L_x, L_y, L_z)$ the amplitudes of the sinusoids; $P_x = P_y = P_z = 3m$ the number of periods of the sinusoidal transformation and $(tpx'_j, tpy'_j, tpz'_j)$ the new target points.

\begin{figure}[!t] 
\captionsetup[subfigure]{labelformat=empty}
    \centering
  \subfloat[\label{DS1a}]{%
       \includegraphics[width=0.5\linewidth]{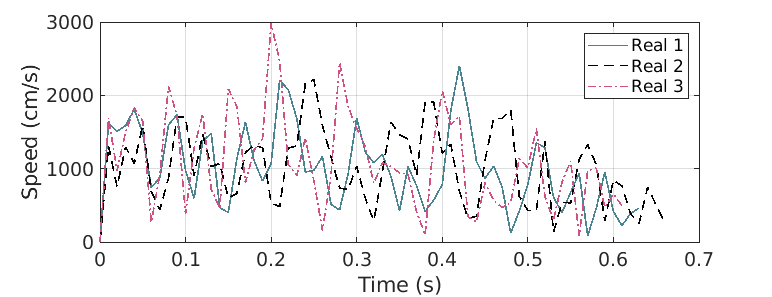}}
    \hfill
  \subfloat[\label{DS1b}]{%
        \includegraphics[width=0.5\linewidth]{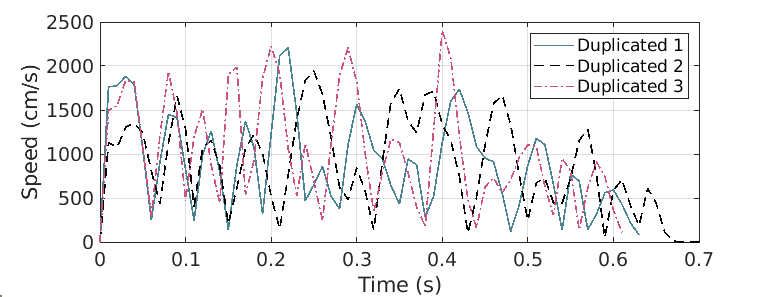}} \\
  \subfloat[\label{DS2a}]{%
       \includegraphics[width=0.5\linewidth]{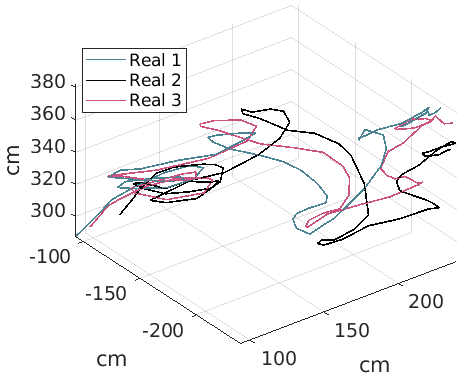}}
    \hfill
  \subfloat[\label{DS2b}]{%
        \includegraphics[width=0.5\linewidth]{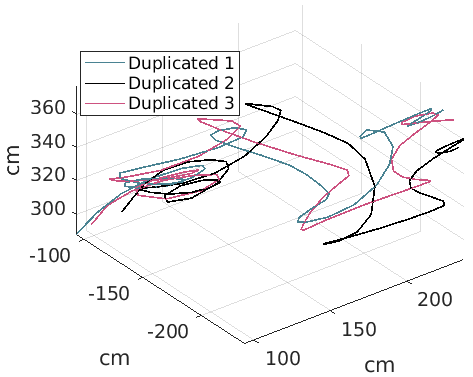}} \\      
  \caption{Examples of 3D real and duplicated signatures with the DS method.}  
  \label{fig:DS} 
\end{figure}

 Next, two affine transformations are applied to the virtual target points: a rotation followed by a displacement, as follows:

\begin{equation}  
\begin{pmatrix}
  {tpx''_j}\\
  {tpy''_j}\\
  {tpz''_j}
 \end{pmatrix} =
\begin{pmatrix}
  {tpx'_j}\\
  {tpy'_j}\\
  {tpz'_j}
 \end{pmatrix}
 \begin{pmatrix}
  1 & 0 & 1 \\
  0 & c(\cdot) & -s(\cdot) \\
  0 & s(\cdot) & c(\cdot)
 \end{pmatrix}
 \begin{pmatrix}
  c(\cdot) & 0 & s(\cdot) \\
  0 & 1 & 0 \\
  -s(\cdot) & 1 & c(\cdot)
 \end{pmatrix}
 \begin{pmatrix}
  c(\cdot) & -s(\cdot) & 0\\
  s(\cdot) & c(\cdot) & 0 \\
  0 & 0 & 1
 \end{pmatrix}
 +
 \begin{pmatrix}
  r\cdot \overline{tpx'}\\
  r\cdot \overline{tpy'}\\
  r\cdot \overline{tpz'}
 \end{pmatrix}
\end{equation}

 \noindent where $c(\cdot)$ or $s(\cdot)$ denote the $\sin$ and $\cos$ operations of $\pi/100 \cdot\mathcal{N}(0; 1)$ on the three rotation matrices, $r \in (0, 0.02)$ is a random value from a uniform distribution used for the displacement effect and $\left(\overline{tpx'}, \overline{tpy'}, \overline{tpz'} \right)$ are the average values of the target point on each axis. According to our observations, a genuine duplicated signature should suffer less distortion than a duplicated forgery.

 Once the new parameters are defined, the signature is reconstructed from $( t'_{0,j}, \mu'_j, \sigma_j^{2'}, \vartheta'_{s,j}, \vartheta'_{e,j}, \phi'_{s,j},$ $tpx''_j, tpy''_j, tpz''_j)$, $j=1,\ldots,N'$, obtaining the online duplicated 3D signature. Three real and three duplicated genuine signatures are illustrated in Figure~\ref{fig:DS} as an example of the DS method.

   \section{Experiments and Results}

 The synthetic 3D signatures were validated in two ways: (i) by generating synthetic 3D signature databases with realistic intra- and inter-person variability and (ii) by confusing humans to judge as real the synthetic signature through a Turing Test. Finally, for a proof of concept, an extension of our synthesizer to 3D on-air writing and 3D gesturing was investigated.

   \subsection{Signature Verification Performance}
   
 We examined whether a machine could see synthetic 3D signatures as real ones. To this end, we conducted signature verification experiments in random and skilled forgeries scenarios, using two publicly available databases:

   \begin{enumerate}
    \item     3DIIT Signatures, consisting of 1600 air-written signatures by eighty individuals, recorded using a Leap motion at 60 Hz \cite{behera2018analysis}.
    
    \item Deep3DSigAir, consisting of signatures from forty users: 10 for training, 10 for testing, and 25 forgeries acquired with Intel's creative senz3D depth camera at a sampling rate of 60Hz \cite{malik2020deepairsig}.
   \end{enumerate}

   Next, we generated two types of synthetic counterpart databases. Firstly, the FS+DS method were used to generate a master and duplicates. The FS method matched the databases mentioned above by replicating their lexical and morphological properties such as the distribution of the number of words, letters, flourishes, signature sizes, intra-person variability, sampling frequency, etc. Secondly, the original temporal information was removed, and the signatures were synthesized with the KS method. The results of both methods were compared to the original databases. All generated databases had the same organization and number of samples as the original ones.

\begin{figure*}[!t]
\captionsetup[subfigure]{labelformat=empty}
    \centering
  \subfloat[ \label{asv1}]{%
       \includegraphics[width=0.33\linewidth]{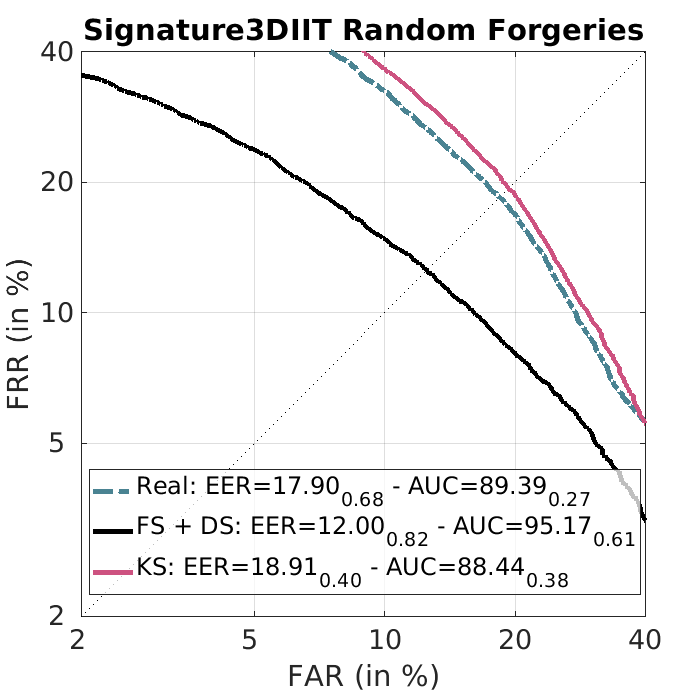}}
    \hfill
  \subfloat[\label{asv2}\textcolor{black}{Results with DTW-based ASV}]{%
        \includegraphics[width=0.33\linewidth]{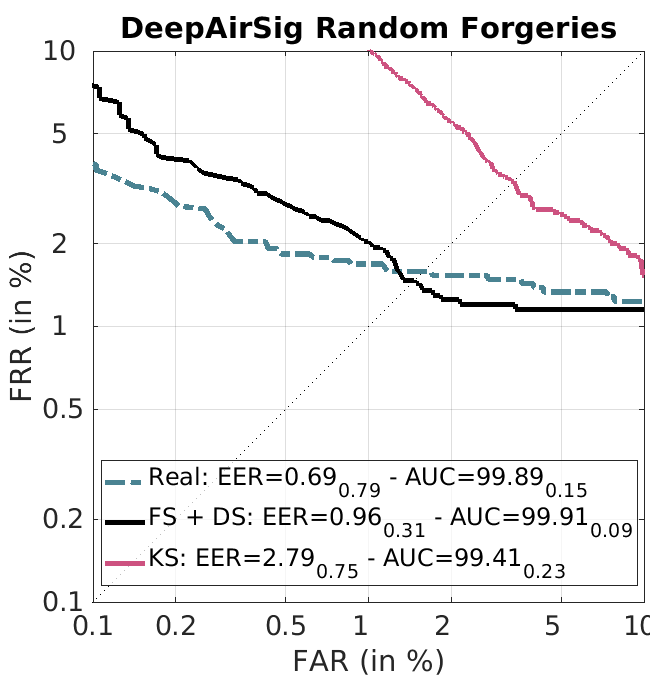}}
   \hfill
  \subfloat[\label{asv3}]{%
        \includegraphics[width=0.33\linewidth]{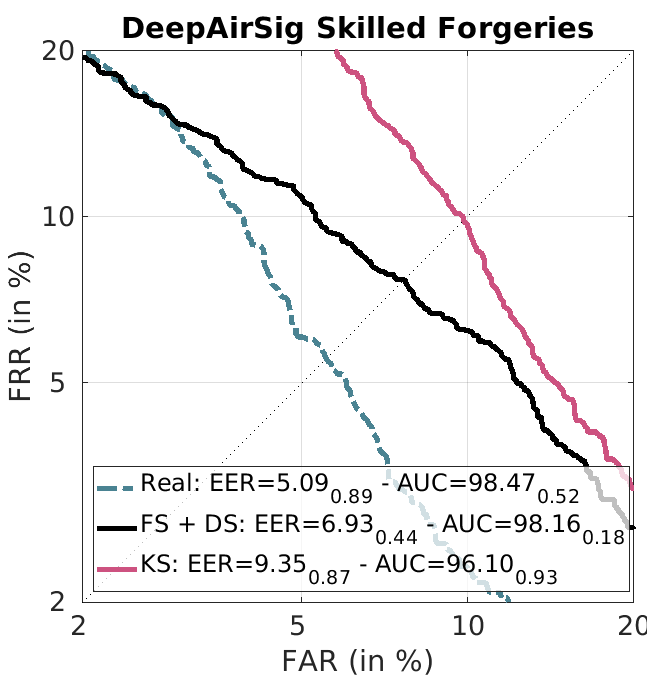}}
        \\
  \subfloat[ \label{asv4}]{%
       \includegraphics[width=0.33\linewidth]{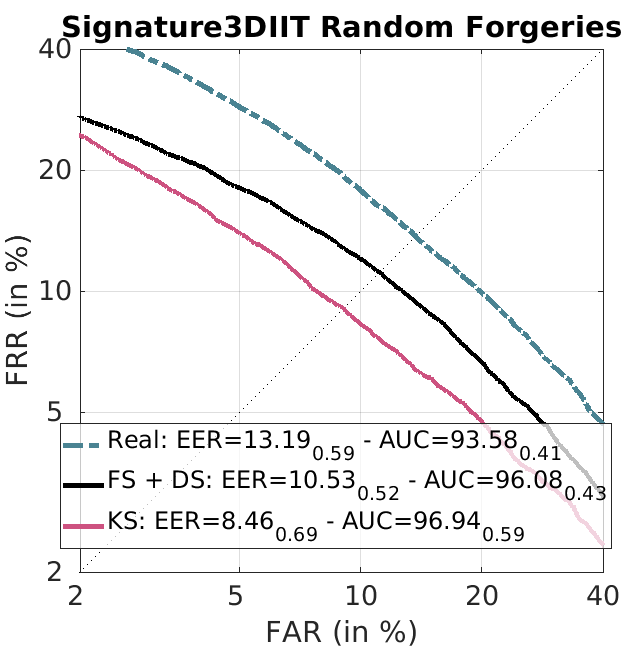}}
    \hfill
  \subfloat[\label{asv5}\textcolor{black}{Results with MAN-based ASV}]{%
        \includegraphics[width=0.33\linewidth]{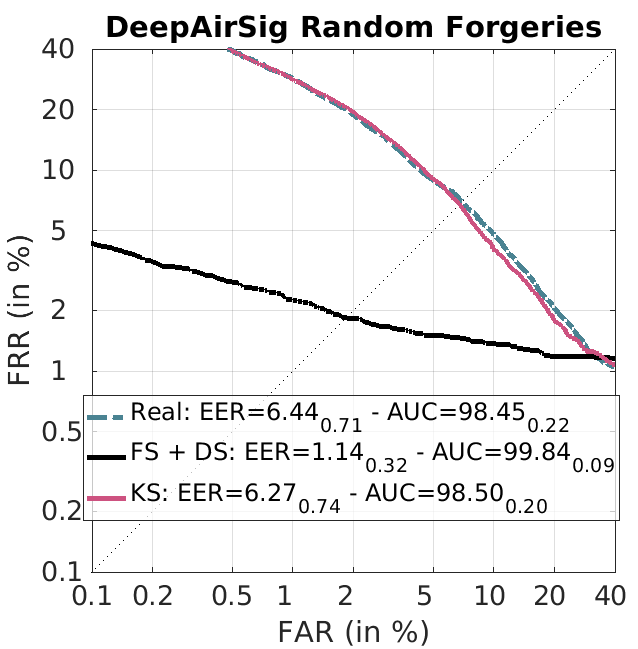}}
   \hfill
  \subfloat[\label{asv6}]{%
        \includegraphics[width=0.33\linewidth]{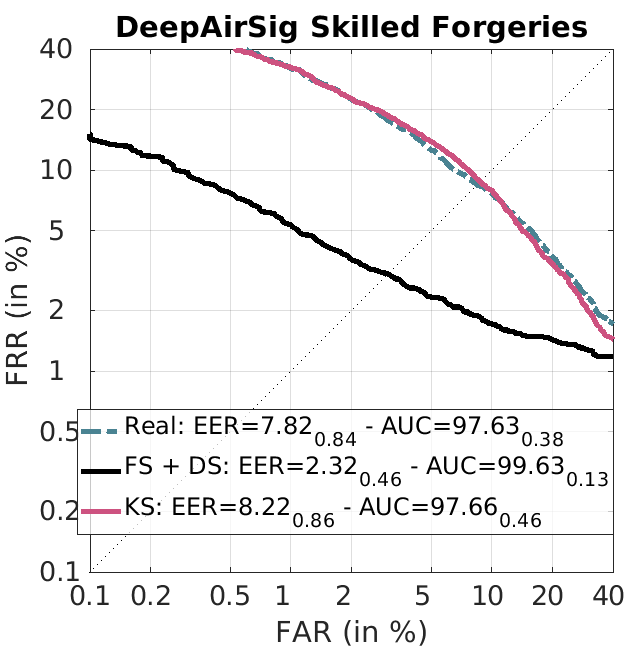}}      
  \caption{Automatic signature verification results with real data and synthetic 3D signatures with FS+DS and KS methods in random and skilled forgeries scenarios \textcolor{black}{for both systems. Above: DTW-based ASV. Below: MAN-based ASV}. \textcolor{black}{(FAR: False Acceptance Rate, FRR: False Rejection Rate, EER: Equal Error Rate, AUC: Area Under Curve).}}
  \label{ASV} 
\end{figure*}

  A DTW algorithm was selected with the 3D trajectory to serve as the ASV, and its first and second derivatives were chosen as features~\cite{parziale2019sm}. Then, we trained the ASV with five random genuine signatures per user. The rest of the signatures were used to test following the standard ASV field benchmark with random and skilled forgeries \cite{diaz2019perspective}. Finally, we repeated the ASV experiments ten times and have presented the average DET plots in Figure~\ref{ASV}.
  
  \textcolor{black}{As results, the synthetic databases displayed a similar behavior as real ones. Specifically, in the case of Signature3DIIT, the generated database with FS+DS seemed to be a bit easier to verify than the real one in the Signature3DIIT. However, an excellent matching was obtained with the KS method in this database. Conversely, the FS+DS was better than the KS in mimicking the Deep3DSigAir in both random and skilled forgeries. For skilled forgeries, our synthetic database seemed more challenging than the real one.}

  \textcolor{black}{	As the synthetic databases were generated to mimic the behavior of real databases using the DTW algorithm, these results may be biased. To test the generalization ability of our synthesizers, we conducted comparative experiments using a different ASV based on the Manhattan (MAN) distance. This verifier was adapted for 3D signature verification from~\cite{sae2014online}. It is based on statistical distributions of the signatures, which are computed in linear time. While the DTW algorithm focuses on temporal intra- and inter-personal variations, the new MAN-based verifier focuses on variations in histograms.}
  
  \textcolor{black}{As a result, similar findings were observed using the MAN-based ASV on the Signature3DIIT database. Specifically, we found a better match between the real and synthetic databases in the case of the Signature3DIIT database. In fact, the standard deviation of the Equal Error Rates (EERs) of the three databases was 2.37 with the MAN method, as compared to 3.73 with the DTW method. In the case of the DeepAirSIg database, the match between the real and synthetic databases was worse with the DTW method in the case of the FS+DS method, but significantly better with the KS method.}

 \textcolor{black}{We can conclude that the match between real and synthetic databases depends on the specific database being mimicked and the verifier used as a reference to adjust the variability in the database. Additionally, the generalization ability also depends on the specific database.} Finally, it is worth pointing out that the proposed synthesizer can be adapted to the tendency of these performances in both random and skilled forgeries, using different databases acquired in several conditions and sensors, such as motion capture suits based on a video and leap motion. Note also that only the variability of the synthesis parameters was modified to show the flexibility of the FS+DS and KS methods.

\begin{figure*}[!b]
       \includegraphics[width=.95\linewidth]{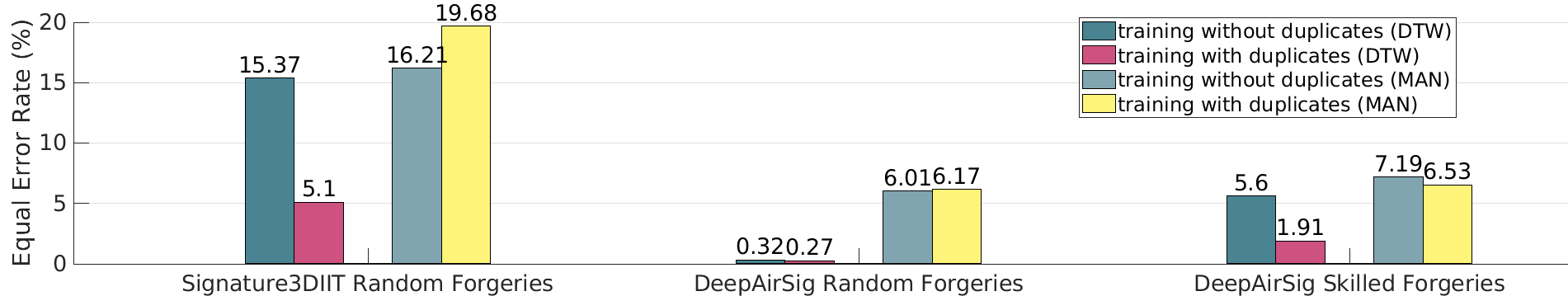}
  \caption{\textcolor{black}{Equal Error Rate improvements by using duplicates in training.}}
  \label{DS_firmas} 
\end{figure*}

  \subsection{Enlarging the Training Signatures with Duplicates}
  
  Of note, the greater the number of signatures used in training, the better the ASV results expected. Following the DS method, we duplicated the training signatures to enlarge the training set. This was a challenging experiment since our duplicates were based on mathematical models and smoother curves. For context, real signatures contain noise errors associated with the acquisition process and user devices, which were not modelled in this work. Accordingly, once the training signatures were parameterized with iDeLog3D, more artificial training samples were generated with the DS method, which tries to capture the real intra-personal variability among subjects. The variation ranges of DS parameters were manually adjusted by trial and error. Thus, by training and testing with the same users and generating ten synthetic duplicated samples per training sample, the average equal error rate (EER) \textcolor{black}{by using the DTW verifier} for 3DIIT Signatures decreased from 15.37\% to 5.10\% (reduction of 66.8\%). The Deep3DAirSig improved from EER=0.32\% to EER=0.27\% for random forgery (reduction of 15.62\%) and from EER=5.06\% to EER=1.91\% (reduction of 62.25\%) in the case of skilled forgeries \textcolor{black}{(see Figure~\ref{DS_firmas})}. \textcolor{black}{In the case of MAN-based ASV, we observed an increase in Deep3DAirSig from EER=16.21\% to EER=19.68\% (a 21.41\% increase). For random forgeries with Deep3DAirSig, we found a constant effect from EER=6.01\% to EER=6.17\% (a 2.66\% increase). However, we also saw an improvement in skilled forgeries from EER=7.19\% to EER=6.53\% (a 9.18\% reduction). Our findings indicate that our duplicates improved the skilled forgeries experiment, which is typically the most challenging. Additionally, for random forgeries, the improvement was more limited when real and synthetic samples were combined with the training of the verifiers. The MAN classifier may be more sensitive to device noise than DTW, making it easier to detect synthetic samples. This suggests the need for further research in modeling this type of noise to improve the synthesizer.} \textcolor{black}{Although this improvement came at the cost of increasing the computational load as the training set was augmented, this experiment shows the usefulness and efficiency  of our duplication method in terms of augmenting datasets. This may be very useful, particularly for developing deep learning systems.}

  \subsection{Turing Test}\label{sub:tt}

 We are aware that a human perception test could quantify the human-like appearance of our synthetic 3D signatures. Thus, a final perceptual Turing test was carried out.
 
 In the first part of the test, 12 short videos with the movement of an end effector were presented to participants, who had to judge if a human or a machine had carried out the action. The results were given in terms of False Machine Rate (FMR), which indicates the percentage of real human samples perceived as machine-made (synthetic), and False Human Rate (FHR), which shows the percentage of synthetic or machine samples perceived as human-made.
 
 In the second part, five videos with two executions of the same movement were displayed to the participants, who had to decide which one was executed by a machine. The results were given in terms of accuracy in determining which was the machine movement.

 The test was carried out through Google forms\footnote{\url{https://forms.gle/bS68BtBPcheo4HNf7}}. We got 369 participants with 6273 answers, with 4428 and 1845 for the first and second part of the test, respectively. 58.84\% of respondents affirmed having computer skills. Regarding age distribution, 4.23\% were under 16, 68.22\% were in the 16-25 range, 19.13\% were in the 26-40 range, and the rest were over 40 years old. On average, the respondents took 5 minutes and 12 seconds to answer the entire test.
   
   Results are given in Table~\ref{tab:usedfeatures}. The FMR was 47.65\% in the first part, and the FHR was 45.67\%. The global accuracy in the second part of the test was 56.32\%. The results show confusion between human and machine movements at nearly 50\%, confirming the human-likeness of the synthesis procedure.

 {\renewcommand{\arraystretch}{0.75}%
\begin{table}
\caption{Global and database Turing test results}
\begin{center}
\begin{tabular}{|l|c|c|c|}
\hline
& Part I & Part I & Part II \\  
& FMR & FHR & Accuracy \\ \hline \hline
Signature3DIIT&
42.24\%&
47.29\%&
56.32\% \\ \hline
Deep3DAirSig&
53.07\%&
44.04\%&
56.32\% \\ \hline
\bf All databases&
\bf 47.65\%&
\bf 45.67\%&
\bf 56.32\% \\ \hline
\end{tabular}
\label{tab:usedfeatures}
\end{center}
\end{table}
}  

\subsection{Extension to Handwriting and Gestures}

 The Sigma-Lognormal model has demonstrated its ability to synthesize handwriting~\cite{fischer2014neuromuscular} and gestures~\cite{leiva2015gestures} in 2D. Accordingly, we used our synthesizer to generate handwriting and gestures in the air and to enlarge the training set with duplicates.

 For writing in the air, we used the AirWriting database, which includes 100 words written once by five users, with 500 words registered with a Leap motion at 60Hz~\cite{chen2015air}. For gestures, two corpora were selected: (i) the HDM05, which contains 11 synthetic actions performed from 2 to 7 times by ten actors, with a total of 250 samples acquired with a VICON system at 120Hz~\cite{muller2007documentation}, and (ii) the UTKinect, which includes ten synthetic gestures written twice by ten actors recorded by a Kinect camera at 30Hz~\cite{xia2012view}.

 For the FS method, we synthetically emulated the AirWriting dataset with a sequence of two to four uppercase letters defining the corresponding virtual target points. For the gestures, a series of five to ten random points inside a 3D cube with a size of 100 pixels was obtained. Let us consider two consecutive virtual target points: the Sagitta of the circumference was randomly obtained between $d/20$ and $d/5$, with $d$ being the distance between these two points. Accordingly, the starting and ending azimuth and polar angles of the arc of circumferences were calculated. Once the gesture was designed based on the iDeLog3D, we also modified its spatial orientation in 3D.

\begin{figure*}[!t]
\captionsetup[subfigure]{labelformat=empty}
    \centering
  \subfloat[ \label{cmc1}]{%
       \includegraphics[width=0.33\linewidth]{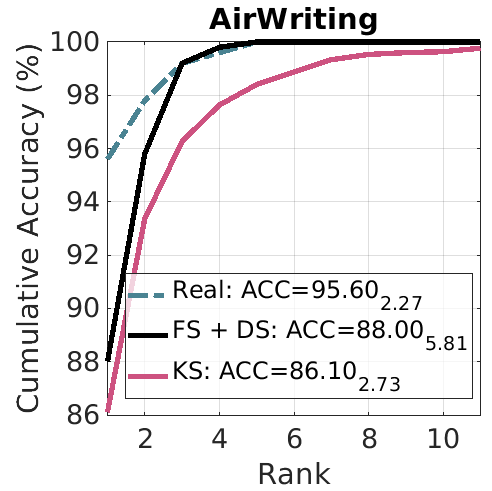}}
    \hfill
  \subfloat[\label{cmc2}]{%
        \includegraphics[width=0.33\linewidth]{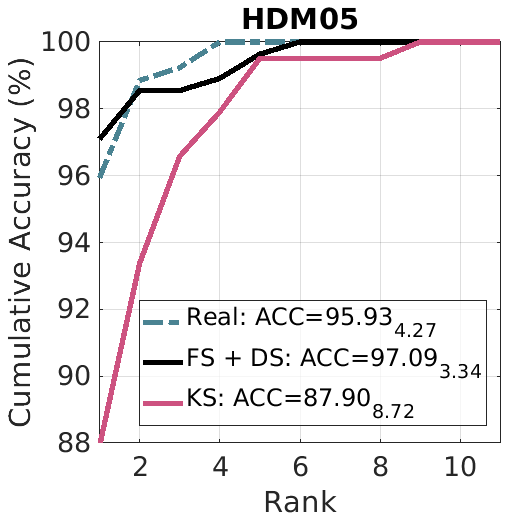}}
   \hfill
  \subfloat[\label{cmc3}]{%
        \includegraphics[width=0.33\linewidth]{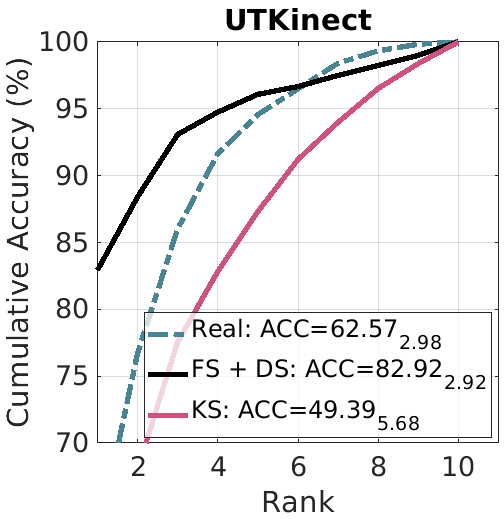}}    
  \caption{Classification results with 3D writing and 3D gestures with real and synthetic samples using the FS+DS and KS methods. \textcolor{black}{(ACC shows the average and standard deviation of the rank-1 recognition rate.)}}
  \label{CMC} 
\end{figure*}

 Next, we reproduced the classification experiments\footnote{As baseline, the classification accuracies reported in~\cite{schindler2018extending, fischer2021modeling} with real databases were AirWriting: Acc=99.0\%, HDM05: Acc=96.4\% and, UTkinect: Acc=94.0\%.} in~\cite{schindler2018extending, fischer2021modeling}, which were also based on a DTW algorithm. The experiments were repeated ten times with real and synthetic databases, and the cumulative match curves are averaged and plotted in Figure~\ref{CMC}. In the AirWriting database, we can see that the FS+DS and KS achieved similar accuracies. However, the FS approximated the real one with ranks greater than three. For gestures, the FS+DS curves were closer to the corresponding real databases than was the KS. This was especially the case in the HDM05, which approximated practically all ranks. In all cases, the KS followed the tendency of real curves, but with a slightly lower performance. In comparison, the FS+DS showed a similar accuracy to the real one in all cases. 

\begin{figure*}
       \includegraphics[width=.9\linewidth]{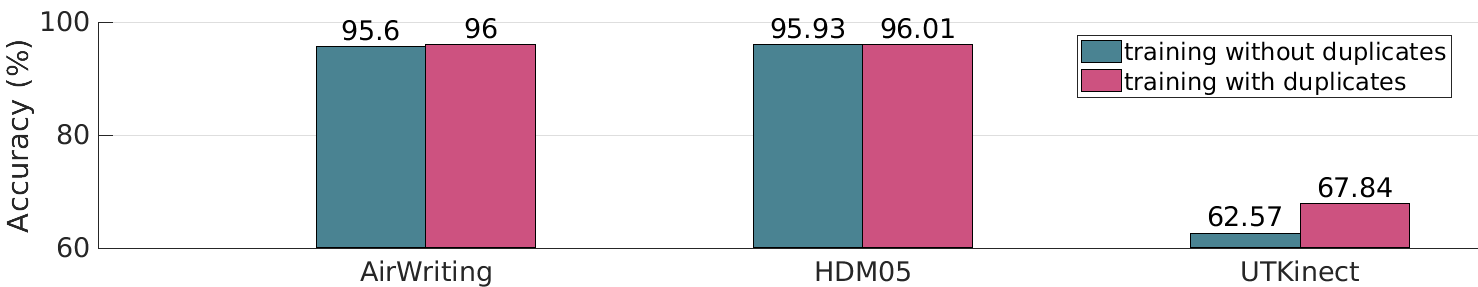}    
  \caption{\textcolor{black}{Accuracy improvements by using duplicates in training.}}
  \label{DS_gestos} 
\end{figure*}

 As duplication is also a relevant stage in synthesis, experiments similar to the signatures experiments were studied. Once again, the training and testing were performed with the same users, and ten synthetic samples per training sample were generated. As a result, the average accuracy for AirWriting increased from 95.60 to 96.00\%; HDM05 improved from 95.93\% to 96.01\%, and UTKinect, from 62.57\% to 67.84\% \textcolor{black}{(see Figure~\ref{DS_gestos})}. These insignificant improvements demonstrated that the duplicator is more effective in signatures, at least when it comes to working with these databases.

 This proof of concept suggests that the proposed methods used to synthesize and duplicate signatures can also be applied to the case of short movements, such as with the characters found in AirWriting and in the 3D gestures. It is worth pointing out that a wide range of sampling frequencies varying from 30Hz up to 125Hz was tested in the experiments, tackling different scenarios. 
  
\section{Conclusion and \textcolor{black}{Discussion}}
 
 This paper focused on synthesizing 3D handwriting signatures following the lognormality principle of human kinematics. Based on a reverse-engineered implementation of iDeLog3D~\cite{idelog3D_2022} analysis, three synthesis methods were proposed: (i) a full synthesis of master 3D on-air signatures, named FS; (ii) the human-like kinematic synthesis of a given spatial on-air trajectory, named KS, and (iii) the synthesis of duplicated 3D on-air signatures, named DS. In addition, the synthesis procedure allowed modifying the level of difficulty of the recognition problem.
 
 The experiments performed herein showed that the FS+DS and KS can generate databases of synthetic movements carried out by artificial users that perform similarly to real ones in verification experiments. Also, it was demonstrated that the DS method could augment real databases for classification problems. Furthermore, a Turing test was carried out to measure the human ability to discriminate between real and synthetic movements. The confusion obtained was close to 50\%, confirming the human likeness of artificial movements in 3D. Finally, an extension of our experiments demonstrated the capacity of our methods to synthesize 3D writing and 3D gestures.

 \textcolor{black}{The literature identifies three main proposals to synthesize signatures: full synthesis, duplication, and analysis of system vulnerabilities. In the case of full synthesis, specimens are generated from scratch~\cite{galbally2012synthetic}. Here, from some statistical distributions, generators were able to produce synthetic signatures and large databases~\cite{ferrer2016behavioral}. One of the main objectives was to design databases that perform similarly to real ones. In the case of duplication, the goal is to use real signatures as seeds to produce other specimens, while keeping the personal intra-variability~\cite{maruyama2020intrapersonal} and boosting the performances~\cite{galbally09ICDARenrollment, diaz2016dynamic}. Furthermore, other works focused on telling human and machine signatures apart~\cite{leiva2021human}. Synthesis-related progress has led to the generation of signatures that are so realistic that their detection by both human verifiers (as indicated in Sect.~\ref{sub:tt} or  \cite{ferrer2013realistic}) and automatic verifiers~\cite{gal2021swagan} is truly challenging.}

 \textcolor{black}{ One aspect needed to improve the realism of our synthetic signatures is their smooth appearance because they are generated by summing Lognormals and circumferences, as is the case in the Sigma-Lognormal model. Furthermore, while retaining the neuromotor background of the Sigma-Lognormal model~\cite{ferrer2018idelog}, it would be interesting to add imperfections found in real handwriting~\cite{leiva2021human}, such as acquisition irregularities of digitizers or ink deposition problems, etc.}
 
 \textcolor{black}{On the other hand, although the realism of synthetic signatures can always be improved, current results open the door to misuse and fraud. Indeed, these procedures allow to generate duplicates that can successfully fool an automatic signature verifier. Moreover, brute force attacks or more sophisticated like hill-climbing techniques~\cite{barrero11BioID} represent a digital threat to signature verification based systems.}
 
 The synthesizer, databases generated in this paper, and Turing Test videos are available for interested readers as supplementary material, which may be helpful in many applications such as human-computer interfaces, medicine, education, games, etc. One of the interests in designing challenging synthetic databases is in the context of forgery generation. The capacity to simulate different degrees of skills allows the signature verification community to develop more precise ASVs. We thus hope that this paper's models and tools will help researchers develop 3D human-like kinematics.

 \section*{Declaration of competing interest}
 
 The authors declare that they have no known competing financial interest or personal relationship that could have appeared to influence the work reported in this paper.

\section*{Acknowledgments}

 This study was funded by the Spanish government's MIMECO PID2019-109099RB-C41 re-search project and European Union FEDER program/funds, the CajaCanaria and la Caixa bank grant 2019SP19, and NSERC grant RGPIN-2015-06409.

\appendix

\section{3D Sigma-Lognormal model}
\label{idelog3D}

The Kinematic Theory of Rapid Human Movements and its associated Sigma-Lognormal model~\cite{plamondon2014recent}  provide a velocity $\vec{v}_o(t)$ modelling regardless of the dimension of the movement. Two different methods have been proposed herein to calculate the $\Sigma\Lambda$ - parameters in 3D: ScriptStudio~\cite{fischer2021modeling}, which optimizes the velocity, and iDeLog (iterative Decomposition in Lognormals)~\cite{idelog3D_2022}, which simultaneously optimizes the speed and the trajectory. For the tridimensional case, in this work, let $\left(x_o(t), y_o(t), z_o(t) \right)^T$ be the observed trajectory of a 3D human movement whose vectorial velocity and module are defined by:

\begin{equation}
 \vec{v}_o(t) = \left(v_{ox}(t), v_{oy}(t), v_{oz}(t) \right) = \left(dx_o(t)/dt, dy_o(t)/dt, dz_o(t)/dt \right)
\end{equation}

\begin{equation}
 v_o(t) = |\vec{v}_o(t)|  = \sqrt{v^2_{ox}(t) + v^2_{oy}(t) + v^2_{oz}(t)}
\end{equation}

The velocity $\vec{v}_o(t)$ is decomposed as a sum of temporally overlapped strokes. Each stroke is a spatiotemporal trajectory between two virtual target points with a lognormal velocity profile~\cite{plamondon2014recent}. Thus, the velocity is modelled as a sum of $N$ lognormals as:

\begin{equation}
 \vec{v}_r(t) = \displaystyle \sum_{j=1}^N D_j \begin{bmatrix}
       \sin(\phi_j(t))\cos(\vartheta_j(t))         \\[0.3em]
       \sin(\phi_j(t))\sin(\vartheta_j(t))           \\[0.3em]
       \cos(\phi_j(t))  
     \end{bmatrix} v_j(t)
\end{equation}

Regarding the trajectory, $\phi_j(t)$ and $\vartheta_j(t)$ are the azimuth and polar angles that represent a planar arc of circumference between the two virtual target points corresponding to the $j^{th}$ stroke.

Once the Sigma-Lognormal parameters $\left\lbrace D_j, t_{0,j}, \mu_j, \sigma^2_j, \vartheta_{s,j}, \vartheta_{e,j}, \phi_{s,j}, \phi_{e,j}  \right\rbrace_{j=1}^N$ are obtained, the velocity  $\vec{v}_o(t)$ can be reconstructed as $\vec{v}_r(t)$, and the trajectory recovered as:

\begin{equation}
 \vec{s}_r(t) = \left( x_r(t), y_r(t), z_r(t) \right) = \displaystyle \int_0^t \vec{v}_r(\tau)\,d\tau
\end{equation}

The similarity between the original velocity $v_o(t)$ and the reconstructed velocity $v_r(t)$ is calculated through the Signal-to-Noise Ratio, $SNR_v$~\cite{ferrer2018idelog}.

The reconstruction quality of the trajectory is determined through the Signal-to-Noise Ratio, $SNR_t$  of the 8-connected original and reconstructed trajectories~\cite{ferrer2018idelog}.

\begin{equation}
 SNR_v = 10 \log \left( \dfrac{\displaystyle \int_0^T |\vec{v}_o(t)|^2\,dt}{\displaystyle \int_0^T |\vec{v}_o(t) - \vec{v}_r(t)|^2 \, dt}\right)
\end{equation}

\begin{equation}
 SNR_t = 10 \log \left( \dfrac{\displaystyle \int_0^T |{s}_o(t) - \bar{s}_o|^2\,dt}{\displaystyle \int_0^T |{s}_o(t) - {s}_r(t)|^2 \, dt}\right)
\end{equation}

\noindent where $T$ is the temporal length of the movement, $|\cdot|^2$ is the Euclidean norm~2, $\bar{s}_o$ the time average and, $s_o(t)$ and $s_r(t)$ the module of the observed and reconstructed trajectory, respectively.


\bibliography{main}

\clearpage

%
%
%
%
%
%

\end{document}